\documentclass[10pt,twocolumn,letterpaper]{article}

\usepackage{iccv}              

\definecolor{iccvblue}{rgb}{0.21,0.49,0.74}
\usepackage[pagebackref,breaklinks,colorlinks,allcolors=iccvblue]{hyperref}
\usepackage{booktabs, multirow, tabularx}
\usepackage[accsupp]{axessibility}
\renewcommand{\thefootnote}{\fnsymbol{footnote}}

\def\paperID{8346} 
\def\confName{ICCV}
\def\confYear{2025}

\usepackage[hang,flushmargin]{footmisc}
\newcommand\blfootnote[1]{%
    \begingroup 
    \renewcommand\thefootnote{} \footnote{#1}%
    \addtocounter{footnote}{-1}%
    \endgroup 
}

\title{TextMaster: A Unified Framework for Realistic Text Editing \\via Glyph-Style Dual-Control}

\author{ 
 Zhenyu Yan$^{1*}$, Jian Wang$^{1*\ddag}$, Aoqiang Wang$^{*\dag}$, Yuhan Li$^{2\dag}$, Wenxiang Shang$^{1}$, Ran Lin$^{1}$ \\[2pt] 
$^{1}$Taobao \& Tmall Group of Alibaba  \hspace{1em} $^{2}$Shanghai Jiao Tong University \\
{\tt\small  \{yeeshai.yzy, alex.wj\}@taobao.com, wangaq1999@gmail.com, melodious@sjtu.edu.cn}
}

\begin{document}
\maketitle
\begin{abstract}
In image editing tasks, high-quality text editing capabilities can significantly reduce both human and material resource costs. Existing methods, however, face significant limitations in terms of stroke accuracy for complex text and controllability of generated text styles.
To address these challenges, we propose TextMaster, a solution capable of accurately editing text across various scenarios and image regions, while ensuring proper layout and controllable text style. 
Our method enhances the accuracy and fidelity of text rendering by incorporating high-resolution standard glyph information and applying perceptual loss within the text editing region. Additionally, we leverage an attention mechanism to compute intermediate layer bounding box regression loss for each character, enabling the model to learn text layout across varying contexts. Furthermore, we propose a novel style injection technique that enables controllable style transfer for the injected text. Through comprehensive experiments, we demonstrate the state-of-the-art performance of our method.

\end{abstract}
 \blfootnote{
  *Equal contribution.\\
  \dag Work done during an internship at Alibaba Group.\\
  \ddag Corresponding author.
}
\begin{figure}[htp]
\centering
\includegraphics[width=1\columnwidth]{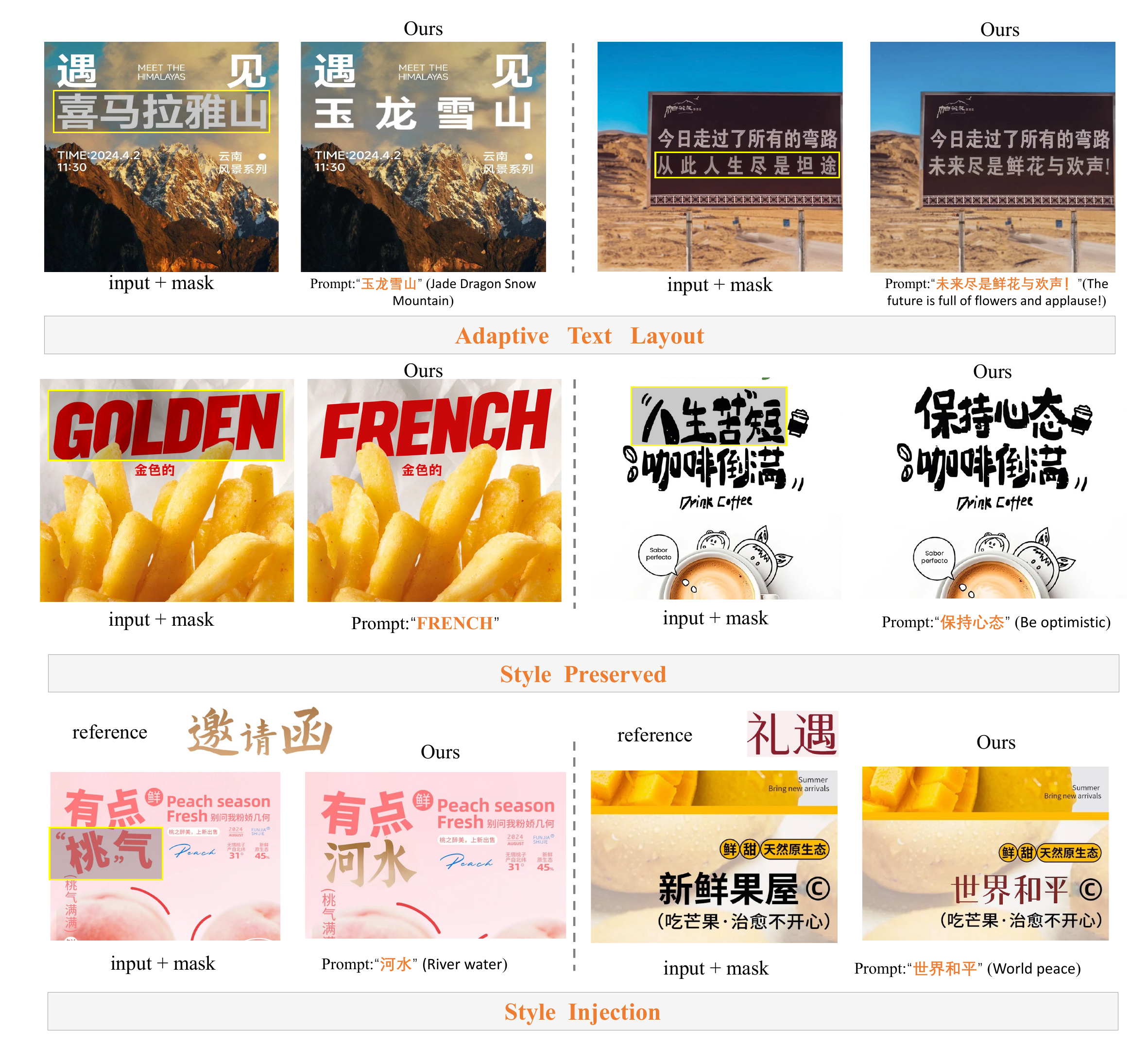}
\caption{The image illustrates the diverse capabilities of our TextMaster, encompassing precise typesetting and layout, consistent style retention, and external style injection.}
\label{fig:intro}
\end{figure}

\begin{figure*}[htp]
\centering
\includegraphics[width=1.8\columnwidth]{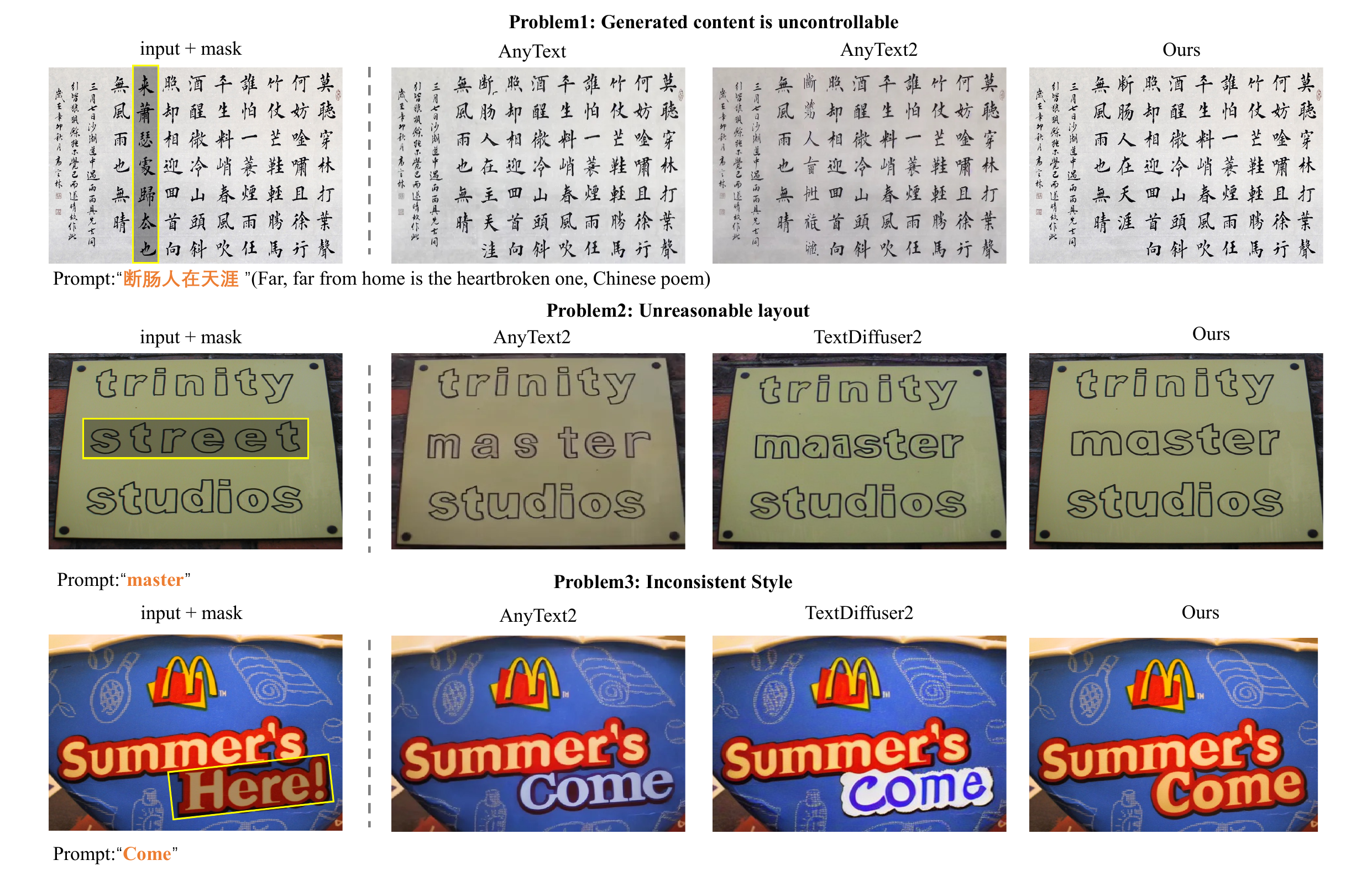}
\caption{The current approach exhibits several deficiencies, including a lack of alignment between the generated text and the intended target text, suboptimal layout organization, and inconsistent in style preservation.}
\label{fig:anytext_compare}
\end{figure*}

\section{Introduction}

Recent advances in generative models have revolutionized image synthesis, with diffusion models surpassing Generative Adversarial Networks (GANs) in quality, diversity, and controllability \cite{ho2020denoising, podell2023sdxl, rombach2022high}. While GAN-based models, such as the original Generative Adversarial Nets \cite{Goodfellow_GAN_NPIS14}, have demonstrated impressive realism, they often suffer from training instability and mode collapse. The advent of diffusion models, such as Denoising Diffusion Probabilistic Models (DDPM) \cite{ho2020denoising} and Latent Diffusion Models (LDM) \cite{rombach2022high}, have led to superior image fidelity and text-image alignment, making them the dominant paradigm in text-to-image (T2I) generation.

To enhance the controllability of T2I generation, researchers have explored methods to inject conditional signals. ControlNet \cite{Zhang_ControlNet_Corr23} enables spatial constraints by conditioning diffusion models on structural priors, while T2I-Adapter \cite{Mou_T2I_Corr23} further enhances control by learning task-specific adapters. Similarly, Composer \cite{Huang_Composer_Corr23} introduces a framework for flexible, composable conditions, allowing intricate constraints in text-guided image synthesis.

Another critical direction involves capturing specific styles and characteristics from reference images. Techniques such as Textual Inversion \cite{Gal_Inversion_ICLR23} and DreamBooth \cite{Ruiz_DreamBooth_Corr22} fine-tune models on personalized concepts, improving fidelity to target attributes. IP-Adapter \cite{ye2023ip-adapter} extends this by efficiently integrating image prompts with textual guidance, while Reference-Efficient \cite{hu_animateanyone} and InstantID \cite{wang2024_instantid} enable identity preservation in personalized generation. Further advancements, such as PhotoMaker \cite{li2023photomaker}, refine identity embeddings to achieve photorealistic results.

The main challenge of text editing lies in generating accurate textual content while minimizing any noticeable traces of editing. To Achieve this requires, the model needs to reconstruct multiple densely packed characters within localized regions, which places substantial demands on its fine-grained modeling abilities. 
Recent methods~\cite{Yang_GlyphControl_Corr23, Ma_glyphdraw_Corr23} have explored glyph control for text editing. Yet, these approaches often suffer from reduced editing accuracy due to the lack of character-level textual guidance. 
On the other hand, effectively disentangling content from style presents another significant hurdle. Due to the highly structured nature of character glyphs, independently extracting style features from a given text image remains particularly difficult. Diffusion-based methods~\cite{Chen_TextDiffuser_Corr23,chen2023diffute,tuo2023anytext} attempt to restore text by leveraging the image context to preserve style. Since style information is primarily derived from unmasked regions, this may lead to inconsistencies in style representation during reconstruction.
Other approaches~\cite{shi2025fonts,ji2023diffste} seek to disentangle text style from glyph structure using manually constructed datasets, where different texts are rendered with various fonts to create style-content pairs for learning. Despite being designed to simulate real-world scenarios, such datasets still fall short in capturing the authenticity and diversity found in real text images.
Furthermore, some methods opt to inject reference styles through text-style image inversion~\cite{zeng2024textctrl, santoso2023dbest}, but this comes at the cost of significantly increased inference overhead.

In response to these challenges, our method introduces further improvements in glyph control by combining character-level text embedding and IoU loss enabling precise regulation of both glyph shape and positioning. For style disentanglement, unlike previous approaches that rely on synthesized data, whose results are often affected by the glyphs in the reference image, we explicitly separate the glyph structure from the style appearance, which is achieved by integrating a style adapter with in-context learning. These innovations enable our model to reconstruct text across deiverse and fully authentic datasets, ultimately delivering robust style disentanglement and superior generalization performance.

By integrating these advanced techniques, the proposed TextMaster framework addresses the critical challenges in text editing, offering unparalleled performance in maintaining style consistency, content accuracy, proper layout, and high-quality visual output. These advancements position TextMaster as a state-of-the-art tool for complex text editing tasks in images \cite{ho2020denoising, dhariwal2021diffusion, saharia2022photorealistic}. Meanwhile, Our approach is not limited to some specific languages, it can be easily extended to other languages as well.

\section{Related Work}

Recent advancements in generative models, such as FLUX \cite{FLUX.1}, DALL·E 3 \cite{DALLE3}, and SD3 \cite{esser2024scaling}, have significantly improved image synthesis quality, achieving high-resolution fidelity and enhanced semantic coherence. However, challenges remain in accurate text rendering within images. Specifically, ensuring precise glyph structure, spatial consistency, and stylistic control remains a complex task.

\noindent \textbf{Accurate Glyph and Spatial Control.} Several studies have addressed the challenges of generating visually coherent and spatially accurate text. GlyphDraw \cite{Ma_glyphdraw_Corr23} proposes a word-level synthesis approach to ensure coherent Chinese character rendering. GlyphControl \cite{Yang_GlyphControl_Corr23} introduces glyph-conditioned diffusion to enhance character structure control, while TextDiffuser \cite{Chen_TextDiffuser_Corr23} treats text rendering as a diffusion-based painting task. Recent advancements, such as AnyText \cite{tuo2023anytext}, extend this to multilingual text generation. Further improvements, such as TextDiffuser-2 \cite{Chen_TextDiffuser2}, leverage large language models for refined text placement. Additionally, models like Glyph-ByT5 \cite{glyphbyt5} and its successor Glyph-ByT5-v2 \cite{glyphbyt5-v2} introduce customized text encoders for multilingual and aesthetic text rendering, improving both accuracy and flexibility. Moreover, GlyphDraw2 \cite{glyphdraw2} combines diffusion models with large-scale language modeling to generate complex text-based posters.

\noindent \textbf{Controllable Text Style and Appearance.} Separating textual content from stylistic attributes is a longstanding challenge. DiffSTE \cite{ji2023diffste} improves text editing by leveraging dual encoders trained on extensive font datasets, allowing font-specific style preservation during synthesis. However, inference still requires explicit font name input. AnyText2 \cite{tuo2024anytext2} enhances controllability by embedding RGB-based color attributes, but complex styles such as gradients, borders, and shadows remain difficult to represent effectively.

\begin{figure*}[ht]
\centering
\includegraphics[width=1.8\columnwidth]{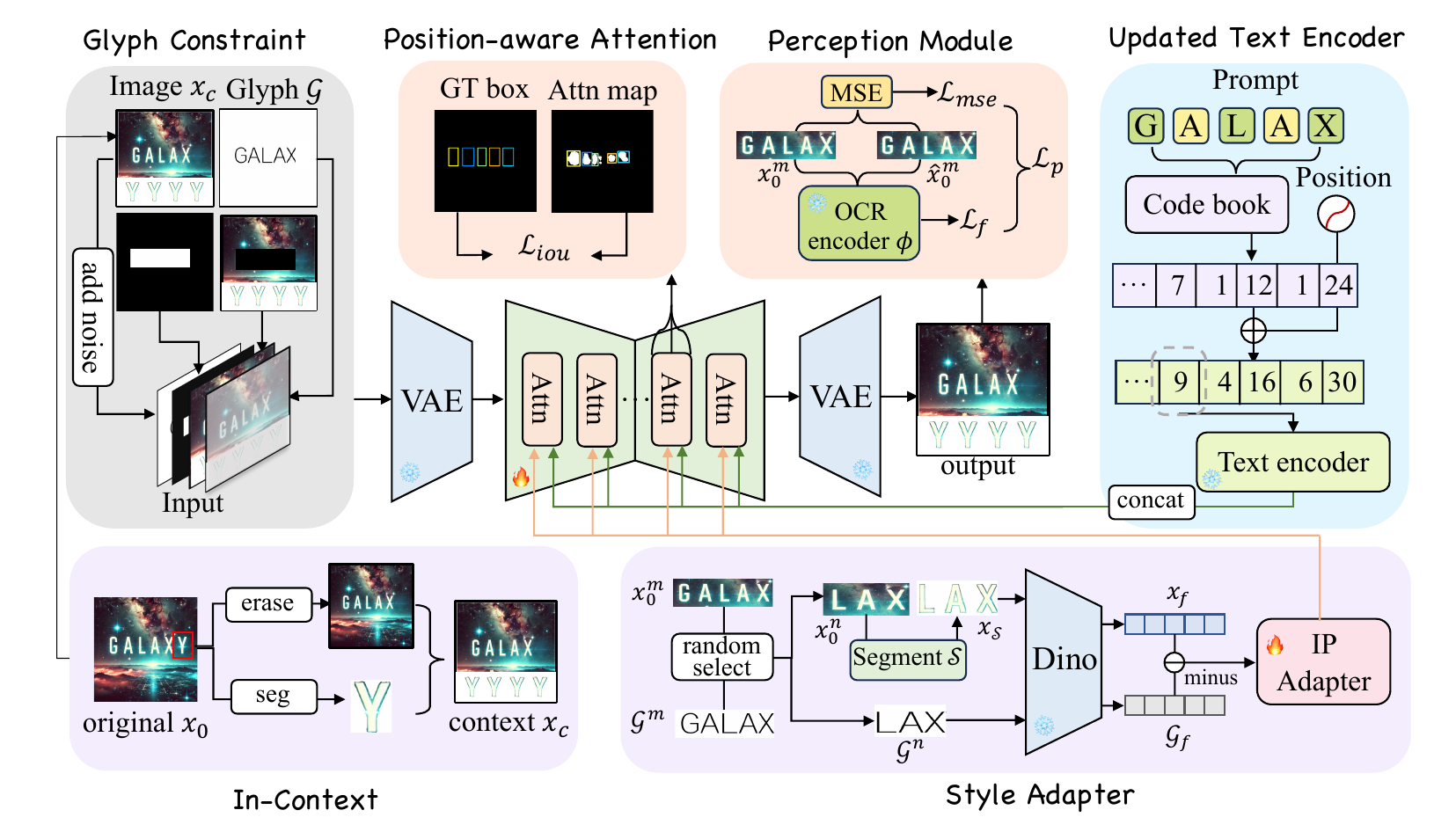}
\caption{The framework of TextMaster consists of the following components: the glyph control module, comprising the glyph constraint, update text encoder, and perception module; the layout control is achieved via position-aware attention; and the style control module encompasses the style-adapter and in-context reference.}
\label{fig:TextMaster}
\end{figure*}

\section{Method}
Text editing tasks can be deconstructed into two aspects: glyphs and styles. As shown in the figure \ref{fig:TextMaster}, to achieve precise editing control, we propose Glyph-Style Dual-Control. Glyph control involves maintaining structural integrity of characters through glyph constraints, and achieving cohesive text flow via adaptive typography layouts. To maintain a consistent style with the original text, we introduce style control, which includes adjustments for typeface and color, allowing for accurate referencing and injection of text styles. This dual mechanism ensures precise character rendering, harmonized layout composition, and versatile style referencing capabilities.

\subsection{Preliminary}
Our proposed method builds upon Stable Diffusion ~\cite{rombach2021ldm} Inpainting Models. Given an image $x_0$, a binary mask $m$ specifying the editable region and a textual prompt $P$ describing the target text to generate, the system first encodes both the original image $x_0$ and its masked image $x_0\odot (1 - m)$ into latent space using a Variational Autoencoder (VAE)~\cite{Diederik_VAE_ICLR14}, where $\odot$ denotes element-wise multiplication. Then a conditional latent diffusion model performs sequential denoising with a UNet in the latent space. UNet takes as inputs the noise latent variable $\hat{z}_t$, downsampled mask $m'$, masked image latent $z_{Im}$, timestep $t$, and embedding $e_t$ of the textual prompt $P$ from a pre-trained CLIP~\cite{radford2021clip} model, and predicts the noise $\epsilon_{\theta}(\hat{z}_t, z_{Im}, m', t, e_t)$. The training of diffusion models resolves around optimizing the denoiser network $\epsilon_{\theta}$ to minimize the noise prediction loss:
\begin{equation}
\mathcal{L}_{\text{noise}}= \left\| \epsilon - \epsilon_{\theta}(\hat{z}_t, z_{Im}, m', t, e_t) \right\|_2^2
\end{equation}

\subsection{Glyph control}

\subsubsection{Glyph Constraint}

The accurate rendering of glyphs is critical in text editing tasks. To ensure proper formation of character strokes, we have implemented multi-faceted constraints including reference standard character images, precise textual encoding representations, and perceptual loss mechanisms.

\noindent \textbf{Glyph Guidance.} 
Previous work ~\cite{Ma_glyphdraw_Corr23,yang2024glyphcontrol,tuo2023anytext} has demonstrated the effective guidance provided by standard font reference images for text content generation. However, these methods design glyph images with per-char alignment positions to accommodate the generation of arbitrary shape masks. Unfortunately, the shape of these masks may unintentionally reveal textual details, such as character count and layout (see Fig.\ref{fig:anytext_compare} problem 1). Therefore, we calculate the outer rectangle of the edit area and inject glyph information by field alignment. The standard quadrilateral mask serves as a better foundation for glyph representation, as a well-coupled relationship between glyphs and layout fosters harmonious content generation.

In detail, we first obtain the bounding box of the text area to be injected and generate the corresponding glyph in this area with standard font in appropriate size. Next, we convert the generated glyph image $\mathcal{I}_g$ into latent space variables $l_g$, using the VAE~\cite{Diederik_VAE_ICLR14} model. Finally, we concatenate these latent space variables with the latent representation of the image along the channel dimension to ensure that the generated image visually aligns with the original input image while clearly retaining and conveying the glyph information.

\noindent \textbf{Token-separated Text Encoder.}
The default text encoder of SDXL~\cite{podell2023sdxl} is not sufficient for representing Chinese semantics. To address this limitations, we employ the ChatGLM~\cite{du2021glm} text encoder to support both Chinese and English characters. ChatGLM employs a dual-stream self-attention architecture: one stream processes positional information to capture text sequence and structure, while the other handles content information to understand semantic relationships.

In text editing tasks, semantic relationships can adversely affect generation outcomes, as character embeddings may differ based on context. To reduce this interference, we adopt a specialized encoding method. First, we tokenize the prompt $P = \{c_1, c_2, \cdots, c_K\}$ by treating each character $c_k \in P$ as an individual token, preventing cross-semantic influence. We then sort and concatenate these tokens according to their original positions, using zero padding to ensure consistent sequence lengths. Finally, we use a text encoder $\mathcal{E}_t$ to encode the characters token by token, incorporating position information $p_k$, and generate the corresponding mask encoding $m_k$ for each token $c_k$. Each character is encoded as follows:
\begin{equation}
e_k = \mathcal{E}_t(c_k + pos_k + m_k),
\end{equation}
where $k \in \{1,2,\cdots, K\}$. We obtain the final text conditional features by concatenating the character encodings:
\begin{equation}
e_t = [e_1;e_2;\cdots;e_K].
\end{equation}
This approach maintains uniform character embeddings across contexts while preserving positional information, ultimately improving the generation quality in text editing tasks.

\noindent \textbf{Perception Module.} 
We propose a method based on perceptual loss to maintain the consistency of glyph structures. By calculating the $L_2$ loss of the Optical Character Recognition (OCR) features in the edited region between the reconstructed image $\hat{x}_0$ and the original image ${x_0}$, we impose constraints on the generation of glyph structures.

Specifically, by denoising the noise prediction $\epsilon_t$ from the denoising network's output, we can derive an estimate $\hat{z}_0$ from the noise vector $z_t$ as follows: $\hat{z}_0 = \frac{1}{\sqrt{\bar{\alpha}_t}}z_t + \frac{\sqrt{1-\bar{\alpha}_t}}{\sqrt{\bar{\alpha}_t}}\epsilon_t$, where $\epsilon_t \sim \mathcal{N}(\bold{0},\bold{I})$, and $\alpha_t$ denotes the variance scheduling of the Gaussian noise. Furthermore, we utilize a VAE decoder to obtain an approximate reconstruction estimate $\hat{x}_0$. Based on the mask $m$ of the edited region, we crop the original image $x_0$ and the reconstructed image $\hat{x}_0$, resulting in the corresponding images of the edited area $x_0^m$ and $\hat{x}_0^m$, respectively. We employ the PPOCR~\cite{li2022ppocrv3} model as the glyph encoder $\phi(\cdot)$ and extract features $f^m = \phi(x_0^m)$ and $\hat{f}^m = \phi(\hat{x}_0^m)$ from the fully connected layer to compute the L2 loss of the glyph features between the original and predicted images. 

Moreover, we found it necessary to ensure a natural transition between the text and the surrounding background. Therefore, we further introduce pixel-based Mean Squared Error (MSE) loss to ensure the accuracy of the text generation area. The resulting perceptual loss is as follows:

\begin{equation}
\mathcal{L}_{p} = \left\| f^m - \hat{f}^m \parallel^2_2 + \parallel x_0^m - \hat{x}_0^m \right\|_2.
\end{equation}

Perceptual loss injects a stronger supervisory signal by constraining the reconstruction of the editing region, ensuring the correct presentation of the text.

\begin{figure}[htp]
    \centering
    \includegraphics[width=8cm]{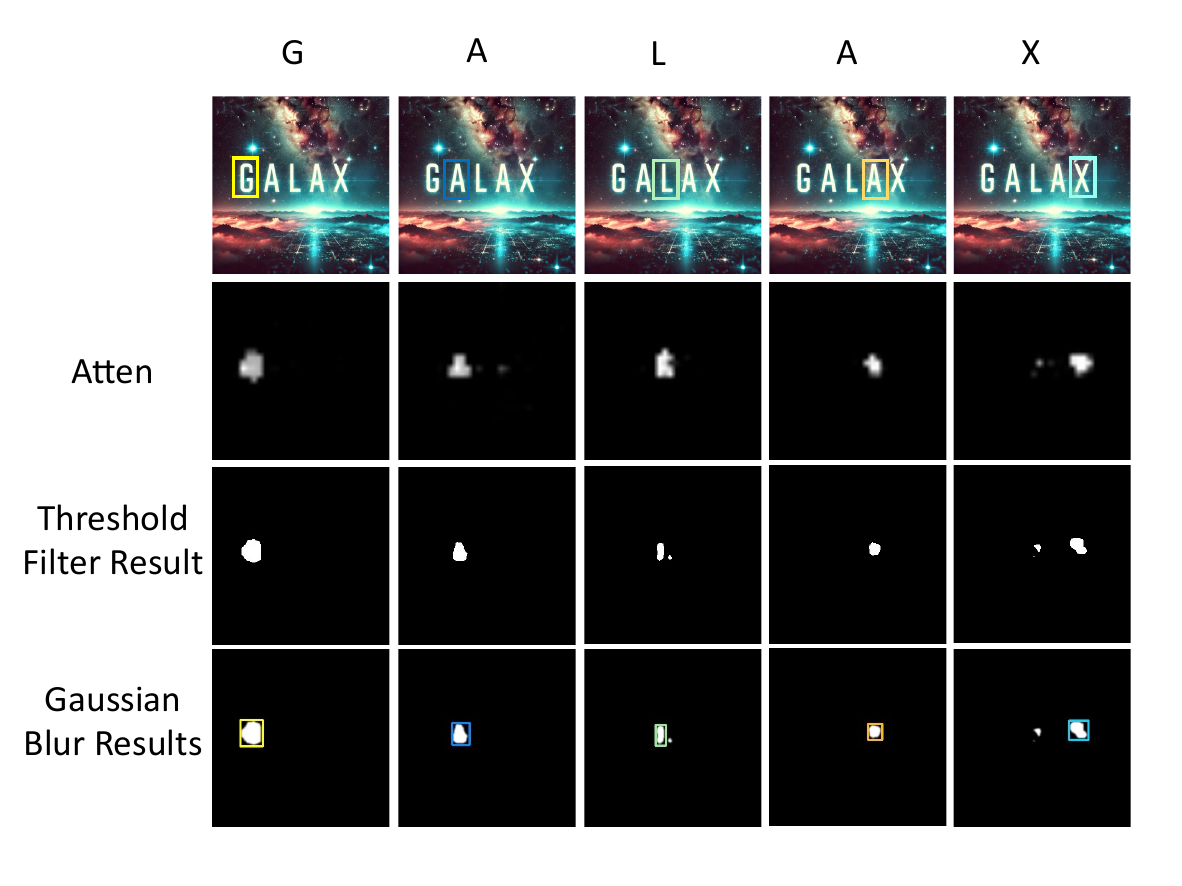}
    \caption{The attention response map processing flow: Atten represent the the top chosen attention response maps for each token.}
    \label{fig:atten_iou}
\end{figure}

\subsubsection{Adaptive Typography Layout}

\noindent \textbf{Position-aware Attention.} 
We expect the model to exhibit accurate positional responses for each character within the masked regions, thereby generating text sequences with proper typographic arrangement. As revealed in prior work ~\cite{wang2024instantstyle}, different cross-attention blocks demonstrate distinct attention orientations, with certain layers exhibiting stronger responses to layout patterns. As illustrated in Fig. \ref{fig:atten_iou}, our analysis of text generation tasks shows that specific attention layers demonstrate pronounced perception of character positioning. To optimize layout learning, we calculate the IoU overlap ratio between binarized attention maps and ground-truth character bounding boxes, subsequently selecting the top three layers exhibiting the strongest responses for layout adaptation. This design facilitates improved intra-paragraph layout learning, as illustrated in Fig. \ref{fig:anytext_compare} problem 2. Specifically, we perform the following steps:

Compute attention scores $\mathbf{A}$ using the query matrix $\mathbf{Q}$ and key matrix $\mathbf{K}$:
\begin{equation}
\mathbf{A} = \frac{\mathbf{Q} \mathbf{K}^\top}{\sqrt{d_k}}, \quad \mathbf{W} = \text{softmax}(\mathbf{A}).
\end{equation}

Thresholding:  
Generate a binary image $\mathbf{T}$ by applying a threshold $T = 125$ to the averaged attention map:
\begin{equation}
\mathbf{T}(i, j) = 
\begin{cases} 
255 & \text{if } \mathbf{\bar{A}}(i, j) > T \\
0 & \text{otherwise}
\end{cases}
\end{equation}

Gaussian Blurring:
Apply Gaussian blurring to the thresholded image $\mathbf{T}$ using a Gaussian kernel $\mathbf{W}(k, l)$:
\begin{equation}
\mathbf{G}(i, j) = \sum_{k=-K}^{K} \sum_{l=-L}^{L} \mathbf{W}(k, l) \cdot \mathbf{T}(i+k, j+l).
\end{equation}

Complete Intersection over Union (CIOU) Loss: 
Calculate the CIOU loss between the derived bounding box $\mathbf{B}_d$ and the ground truth bounding box $\mathbf{B}_g$:
\begin{equation}
\mathcal{L}_{\text{iou}} = 1 - \left( \text{IoU} - \frac{\rho^2(\mathbf{B}_d, \mathbf{B}_g)}{c^2} - \alpha v \right).
\end{equation}
Where:
\begin{itemize}
    \item $\text{IoU}$: Intersection over Union of the two bounding boxes.
    \item $\rho(\mathbf{B}_d, \mathbf{B}_g)$: Euclidean distance between the centers.
    \item $c$: Diagonal length of the smallest enclosing box covering both bounding boxes.
    \item $\alpha$: Trade-off parameter.
    \item $v$: Measure of aspect ratio consistency.
\end{itemize}

\begin{figure}[htp]
\centering
\includegraphics[width=\columnwidth]{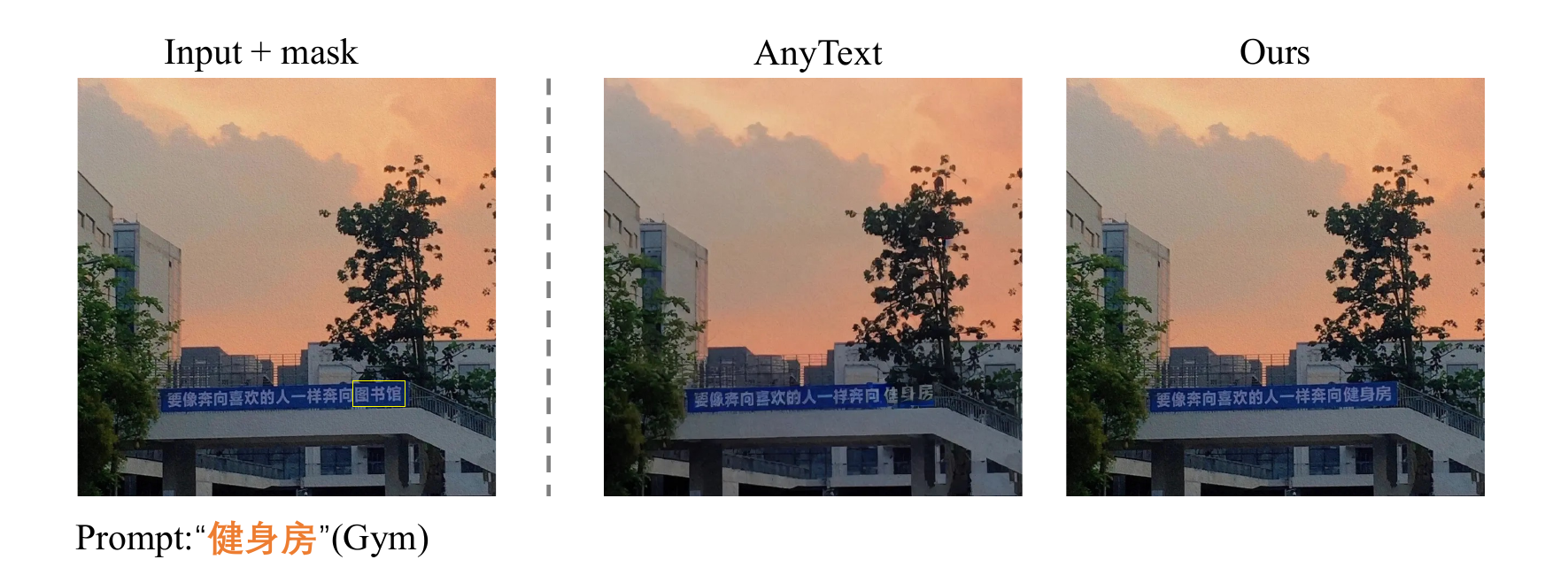}
\caption{The text modification area depicted in the image is merely 30 pixels in height; TextMaster is still capable of executing high-quality edits within this constrained space.}
\label{fig:tiny_compare}
\end{figure}

\noindent \textbf{Adaptive Crop and Mask Strategy.} 
Our experiments revealed that VAE-encoded latent representations often degrade glyph structures in extremely small text. To improve multi-scale generalization, we developed an Adaptive Crop strategy that dynamically adjusts text regions. In our experimental setup, we configured the threshold value at 9 characters. For text segments shorter than this threshold, we performed center-aligned cropping on the images and scaled them to the proportional size corresponding to a 9-character width. When the text length exceeded the threshold, we expanded the cropped region at a 1/10 scaling ratio. As demonstrated in Fig.~\ref{fig:tiny_compare}, this threshold-triggered spatial adaptation significantly improves background restoration and visual coherence.

Additionally, to address uncontrollable text generation caused by mask size leakage (Fig.~\ref{fig:anytext_compare} problem 1) and mitigate layout issues from over-reliance on precise bbox dimensions, we propose an adaptive mask augmentation technique. By randomly expanding masked regions around text, this approach enhances robustness to bbox variations while improving the model's capability to generate visually coherent text layouts.

\subsection{Style Injection}

Current methods prioritize text accuracy and layout over style consistency. Preserving original text styles enhances visual harmony with image contexts, while integrating external references broadens editing versatility. Although recent works~\cite{tuo2024anytext2, ji2023diffste}  employ style control via datasets artificially constructed from font libraries (conditioned on basic attributes like color and typeface), such synthetic datasets often mismatch real-world distributions and provide coarse-grained control, failing to reproduce complex styles like gradient colors, outlined text, or 3D effects.(Fig.~\ref{fig:anytext_compare} problem 3) To overcome these limitations, we propose a Style-Adapter module with In-Context Learning for precise style manipulation.

\subsubsection{Style-Adapter}
In style transfer tasks, decoupling style from content is a critical challenge. Text inherently carries strong glyph information, and the segmentation boundary information carried by the style often disrupts the generation results. InstanceStyle~\cite{wang2024instantstyle} decouples image objects and their semantic content. Inspired by this, we aim to force the model to decouple font style from glyph content.

First, to enable the model to extract more general style information rather than strictly aligning with the segmentation boundaries of the generated text, we randomly select $n, n\in \{1,2,\ldots, K\}$ words from the edit region of the original image $x_0$ and the glyph $\mathcal{G}$, resulting in cropped images $x_0^n$ and $\mathcal{G}^n$. We further apply a text segmentation model $\mathcal{S}$ to separate the font style from the background in $x_0^n$, yielding a segmented font style image $x_{\mathcal{S}} = \mathcal{S}(x_0^n)$.

By employing Dinov2~\cite{oquab2023dinov2} as an image feature extractor $f$, we obtain the features containing glyph and style information $x_{f} = f(x_{\mathcal{S}})$ and the content features $\mathcal{G}_f = f(\mathcal{G}^n)$. We then subtract the extracted feature vectors to obtain content-decoupled style information $\mathcal{F}_\mathcal{S} = x_f - \mathcal{G}_f$, which is injected into the cross-attention layers of the UNet using IP-Adapter~\cite{ye2023ip-adapter}. By jointly injecting image style features and textual prompts as conditional information into the UNet attention block, we achieve interactive fusion through cross-attention mechanisms with the edit image. Furthermore, since image style injection only requires modifications to masked regions, we restrict the style's response regions to text editing areas by computing the intersection between IPA attention maps and the mask region $m$.

During text style injection, treating content bias as style feature resolves the coupling issue between glyph and typographic styles. This approach fully unleashes the text-image style transfer capabilities in two key aspects: (1) Enabling preservation of original typographic styles when editing full text segments without contextual references, through injecting source image characteristics into modified content. (2) Providing external style injection functionality that allows users to edit images using preferred textual exemplars through our style transfer framework.

\subsubsection{In-Context Reference}

Our proposed Style-Adapter follows this approach to inject textual attributes from reference characters, achieving robust font style reconstruction. However, sometimes chromatic deviations persist due to the model's tendency to derive pixel color information from the contextual surroundings of the original image. 

The model's ability to extract color information from original image contexts led us to implement self-attention mechanisms~\cite{zeng2024catdm, huang2024iclora, zhou2024storydiffusion} to manage detailed stylistic features. Based on this, TextMaster introduces an efficient approach to enhance the model's understanding of stylistic nuances in reference text. As illustrated in Figure \ref{fig:TextMaster}, our framework processes the reference character $c_s$ using a segmentation model $\mathcal{S}$. The resized and positioned $c_s$ is placed at the bottom of the input image, creating modified context image $x_c$, and latent features $z_t^c$. This setup allows the model to learn spatial-semantic relationships between masked regions and the added contextual reference through self-attention.

\begin{equation}
\mathbf{Q}_{\text{self}} = l_\mathbf{Q}(\varphi_{\text{self}}(z_t^c)),  \mathbf{K}_{\text{self}} = l_{\mathbf{K}}(\varphi_{\text{self}}(z_t^c)).
\end{equation}

\begin{equation}
\mathbf{A}_{\text{self}} = \text{Softmax}(\frac{\mathbf{Q}_{\text{self}}\mathbf{K}_{\text{self}}^\top}{\sqrt{d_{\text{self}}}}).
\end{equation}

Here, $l_\mathbf{Q}$ and $l_\mathbf{K}$ denotes the linear projection operation, $\varphi_{\text{self}}$ represents layer normalization, and $\mathbf{A}_{\text{self}}$ corresponds to the attention maps derived from self-attention computation. We term this approach \emph{In-Context Reference}. 

The denoising loss is then formulated as follows:
\begin{equation}
\mathcal{L}_{\text{noise}}= \left\| \epsilon - \epsilon_{\theta}(z_t, m', t, e_t) \right\|_2^2
\end{equation}
Our overall loss consists of denoising loss $\mathcal{L}_{\text{noise}}$, perceptual loss $\mathcal{L}_{\text{p}}$ and CIOU loss $\mathcal{L}_{\text{iou}}$ , This can be expressed as:
\begin{equation}
\mathcal{L} =  \mathcal{L}_{\text{noise}} + \beta \mathcal{L}_{\text{p}} + \gamma \mathcal{L}_{\text{iou}}
\end{equation}
Where $\beta$ and $\gamma$ are coefficients that adjust the relative contributions of these loss terms.

During training iterations, we observed that directly using the target reconstruction text as context reference led to an over-reliance on copying contextual glyphs to masked regions during inference, resulting in character deformation. To mitigate this, we randomly select a character as the contextual reference and completely erase it using the erasure model~\cite{suvorov2021lama}, enabling the model to learn reconstruction of remaining characters while decoupling the generated textual content from the referenced typographic style. This strategy enhances the model's capability for accurate reconstruction of color-detail stylistic features and ensures effective content-style disentanglement characteristics.

\begin{figure*}[htp]
\centering
\includegraphics[width=1.9\columnwidth]{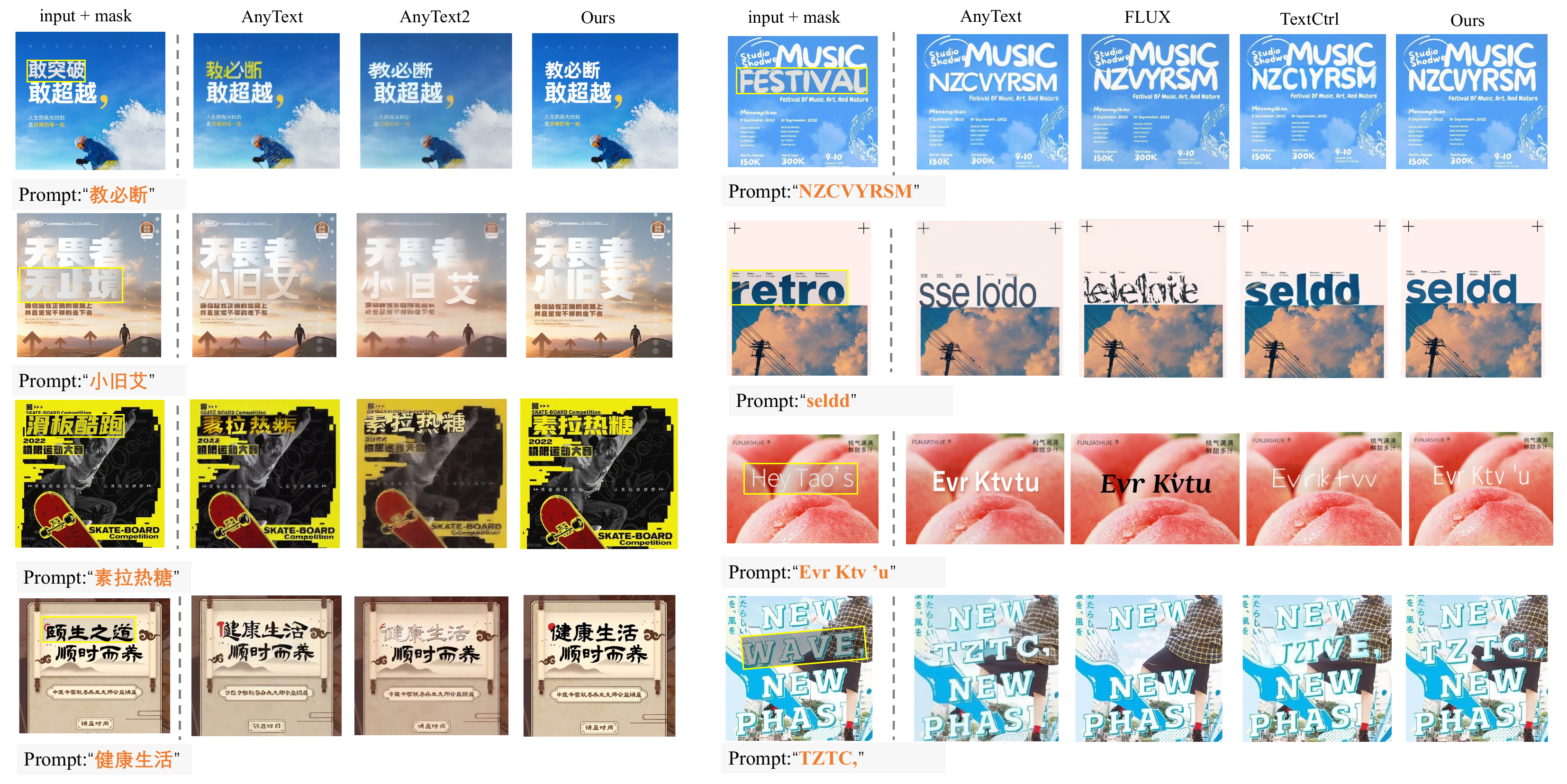}
\caption{Comparison of the visualized results between our method and anytext\cite{tuo2023anytext}, anytext2\cite{tuo2024anytext2},  flux\cite{FLUX.1}, and textctrl\cite{zeng2024textctrl}.}
\label{fig:experiment_compare}
\end{figure*}

\begin{table*}[ht]
\centering
\resizebox{0.9\linewidth}{!}{
\begin{tabular}{cccccccccc}
\hline
\multirow{2}{*}{\begin{tabular}[c]{@{}c@{}}Evaluation\\ Benchmark\end{tabular}} & \multirow{2}{*}{Model}        & \multicolumn{4}{c}{Chinese}          & \multicolumn{4}{c}{English}          \\ \cline{3-10} 
   &         & Acc$\uparrow$  & NED$\uparrow$  & FID$\downarrow$       & LPIPS$\downarrow$& Acc$\uparrow$  & NED$\uparrow$  & FID$\downarrow$       & LPIPS$\downarrow$\\ \hline
\multirow{7}{*}{\begin{tabular}[c]{@{}c@{}}AnyText-\\ Benchmark\end{tabular}}  
   & SDXL         & -          & -          & -           & -           & 0.0857          & 0.1464          & 37.02          & 0.0557         \\
   & FLUX & -          & -           & -           & -           & 0.1594          & 0.3185          & 25.67         & 0.0471          \\
   & TextDiffuser2      & -           & -           & -    & -    & 0.5475          & 0.7951          & 31.54    & 0.1457    \\
   & AnyText-v1.1       & 0.5314          & 0.6437          & 54.81          & 0.1115          & 0.5167         & 0.5877          & 43.97          & 0.1303          \\
   & AnyText2          & 0.6133    & 0.7120    & 30.56    & 0.0536    & 0.6235          & 0.6657          & 28.66         & 0.0662          \\
   & TextCtrl         & -    & -    & -    & -    & 0.8096          & 0.8231          & \textbf{6.01}         & \textbf{0.0234}          \\
   & \textbf{TextMaster}& \textbf{0.9257} & \textbf{0.9483} & \textbf{11.93}         & \textbf{0.0247}          & \textbf{0.9358} & \textbf{0.9534} & 10.35          & 0.0238          \\ \hline
\multirow{7}{*}{\begin{tabular}[c]{@{}c@{}}ICDAR13-\\ Benchmark\end{tabular}}  
& SDXL         & -          & -          & -           & -           & 0.5579          & 0.1024         & 40.52          & 0.0735         \\
   & FLUX & -          & -           & -           & -           & 0.1594          & 0.1373          & 21.64         & 0.0581          \\
   & TextDiffuser2      & -           & -           & -    & -    & 0.5921          & 0.5425          & 40.27    & 0.1537    \\
   & AnyText-v1.1       & -          & -          & -          & -          & 0.5579         & 0.6142          & 34.56          & 0.1536          \\
   & AnyText2          & -    & -    & -    & -    & 0.6765          & 0.6832          & 17.65         & 0.0754         \\
   & TextCtrl         & -    & -    & -    & -    & 0.8055          & 0.8116          & \textbf{10.77}         & 0.0278          \\   
   & \textbf{TextMaster}& - & - & -         & -          & \textbf{0.9100} & \textbf{0.9366} & 10.82          & \textbf{0.0256}          \\ \hline
   
\multirow{7}{*}{\begin{tabular}[c]{@{}c@{}}TextMaster-\\ Benchmark\end{tabular}}   
   & SDXL         & -          & -          & -           & -           & 0.1232          & 0.1675          & 45.34          & 0.0894         \\
   & FLUX & -          & -           & -           & -           & 0.1764          & 0.2837          & 27.32         & 0.0546          \\
   & TextDiffuser2      & -           & -           & -    & -    & 0.3427          & 0.4537          & 19.64    & 0.0846    \\
   & AnyText-v1.1       & 0.5467          & 0.7681          & 27.96          & 0.0934          & 0.4538         & 0.5877          & 18.98          & 0.1303          \\
   & AnyText2          & 0.5784    & 0.6437    & 19.61    & 0.0638    & 0.5499          & 0.6327          & 14.37         & 0.0794          \\
   & TextCtrl         & -    & -    & -    & -    & 0.8223          & 0.8354          & 10.41         & 0.0293          \\
   & \textbf{TextMaster}& \textbf{0.9180} & \textbf{0.9525} & \textbf{9.86}         & \textbf{0.0245}          & \textbf{0.9180} & \textbf{0.9345} & \textbf{9.73}          & \textbf{0.0237}          \\ \hline
   
\end{tabular}}
\caption{Evaluation Results of random text editing on five benchmarks. Accuracy (ACC, NED) is higher the better ($\uparrow$), while FID and LPIPS are lower the better ($\downarrow$).}
\label{table:comparison}
\vspace{-0.4cm}
\end{table*}

\begin{table*}[h]
    \centering
    \begin{tabular}{lc c c c}
        \toprule
        Setting & Acc$\uparrow$ & NED$\uparrow$ & FID$\downarrow$ & LPIPS$\downarrow$ \\
        \midrule
        $\texttt{Base}$ & 0.5098 & 0.5873 & 16.26 & 0.0454  \\
        +\texttt{Token-separated} & 0.6832 & 0.7073 & 14.58 & 0.0417 \\
        +\(\mathcal{L}_{\text{p}}\) & 0.7451 & 0.7638 & 12.95 & 0.0369 \\
        +\(\mathcal{L}_{\text{iou}}\) & 0.7816 & 0.8495 & 13.02 & 0.0351  \\
        +\texttt{in-context} & 0.8162 & 0.8741 & 11.65 & 0.0338  \\
        +\texttt{style adapter} & \textbf{0.8860} & \textbf{0.9025} & \textbf{10.87} & \textbf{0.0286}\\
    \bottomrule
    \end{tabular}
    \caption{Experimental results of different methods.}
    \label{table:ablation}
    \vspace{-0.4cm}
\end{table*}

\section{Experiments}

\subsection{Dataset}

Our model was trained on the AnyWord-3M dataset containing Chinese/English texts, combining LAION~\cite{schuhmann2021laion} (p1/p2) and Wukong \cite{gu2022wukong} (1-2/5) subsets. Given the limitations of existing open-source datasets in both image resolution and diversity, we publicly crawled 100,000 of artistic style posters from the web to enhance the model's style infusion capability during training.
Specially, we employed an OCR character detection model to generate single-character bounding boxes for precise CIOU loss calculation in character positioning and context reference in style learning. In addition, we used ICDAR13\cite{icdar}, Anytext benchmark\cite{tuo2023anytext}, and the textmaster dataset we built for evaluation.

\subsection{Experimental Details}
Our model was initialized using pre-trained weights from the SDXL\cite{podell2023sdxl} model and pre-trained IPA weights from kolors\cite{kolors}. The training process was conducted in two phases. First step, we train five epochs on the AnyWord-3M dataset using 4 H20 GPUs. Set the learning rate to 1e-4, the weight of perception loss $\beta$ to 0.1, the weight of iou loss $\gamma$ to 0.05, and the noise-offset of 0.02. In second stage, 
we continue training Unet with in-context reference on style poster dataset, with learning rate of 1e-5. Then, we train style-adapter module with freezing UNet parameters first, subsequently jointly optimizing IPA-UNet attention interactions, and ultimately enabling full parameter fine-tuning. During training, the resolution is set to 1024.

\subsection{Comparison Results}
\subsubsection{Quantitative and Qualitative Results}

Following the methodology established in prior research~\cite{podell2023sdxl,FLUX.1,tuo2023anytext,tuo2024anytext2,wang2025dreamtext,zeng2024textctrl}, we conducted a comprehensive evaluation of our model across three distinct datasets using a standardized approach. We divided the metrics into those that assess content accuracy, including OCR content accuracy and text edit distance, and for evaluating style retention, including FID and LPIPS. We ensured a fair comparison among all methods by evaluating accuracy metrics using random text, while assessing style similarity metrics with the original text.
For t2i methods \cite{tuo2023anytext, tuo2024anytext2, podell2023sdxl, FLUX.1}, we supplemented them with image description prompts.  As shown in Table~\ref{table:comparison}, our method outperforms the current state-of-the-art methods across all metrics. We observe that TextCtrl tends to retain the original text during the editing process, which leads to reduced accuracy in content editing. However, this characteristic gives it an advantage in style similarity metrics when reconstructing from the original text.

As shown in Figure \ref{fig:experiment_compare}, we have tested random editing and can see the great advantages of our method for editing arbitrary random content. Also, a large number of qualitative results (see supplementary materials) demonstrate that our model significantly outperforms existing models in both Chinese and English editing effects, particularly in complex Chinese text editing. Moreover, our general editing capability not only maintains accurate content editing but also enhances style retention ability.

\subsection{Ablation Study}

As shown in Table \ref{table:ablation}, we conducted ablation evaluations on 500 carefully selected images that represent various aspects of text. The results show that incorporating glyph information significantly improves the model's performance across all metrics. The perceptual losses and token-separated text encoder contribute to a substantial increase in text generation accuracy and overall image quality. Furthermore, considering the attention IoU loss further enhances the model's ability to accurately locate text in images. 

Meanwhile, the style injection module not only significantly improves the metric of style consistency but also synergistically enhances the accuracy of text editing. In our experiments, we observe that style injection is essential for text generation: in the absence of style reference, the generated arbitrary-style text tends to produce less accurate glyphs, and the resulting text is more difficult to harmoniously blend with its surroundings.

\section{Conclusion}
In this paper, we introduce TextMaster, a groundbreaking method that, for the first time, supports high-accuracy, consistent style text editing with rational layout capabilities. We ensured the generation of glyph structures through the design of glyph images, a token-separating encoder, and a perceptual loss method. We utilized attention IoU loss to promote the model's learning of adaptive layouts.
Additionally, we introduce a style injection method that infuses the style of the modified font into the target font, decoupling semantic entanglement and preserving style information. Extensive qualitative and quantitative experimental results demonstrate that TextMaster outperforms the current state-of-the-art methods.
{
    \small
    \bibliographystyle{ieeenat_fullname}
    \bibliography{main}
}
\clearpage
\section*{Appendix}

\begin{figure}[ht]
  \centering
  \includegraphics[width=1.0\linewidth]{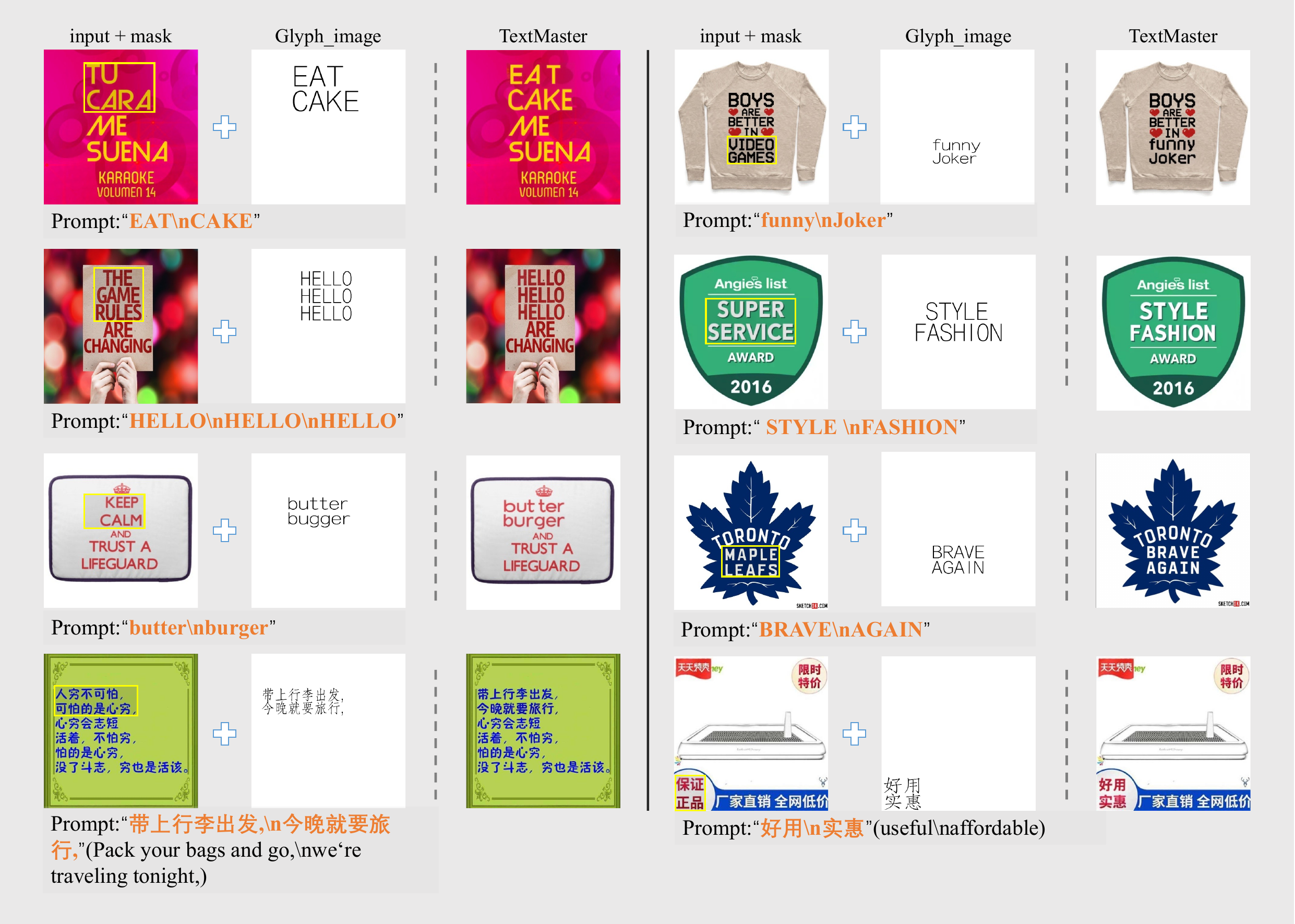}
  \caption{The results of multi-line text editing using \textbf{TextMaster}, with test images sourced from the Laion and WuKong evaluation datasets.}
  \label{fig:duohang}  
\end{figure}

\paragraph{Implementation of Multi-line Text Editing.} As elaborated in the main text, we curated a dataset of 300,000 images containing continuous text segments for training. During the training process, the newline character ("\textbackslash n") was treated as an independent and special token within the prompt. Each line of text content was concatenated using "\textbackslash n", ensuring that the positions of text segments in the glyph images approximately matched their corresponding locations. Guided by the textual information and the injected glyph image data, as shown in Fig.~\ref{fig:duohang}, our method demonstrates a robust capability for precise multi-line text editing.

\begin{figure}[ht]
  \centering
  \includegraphics[width=1.0\linewidth]{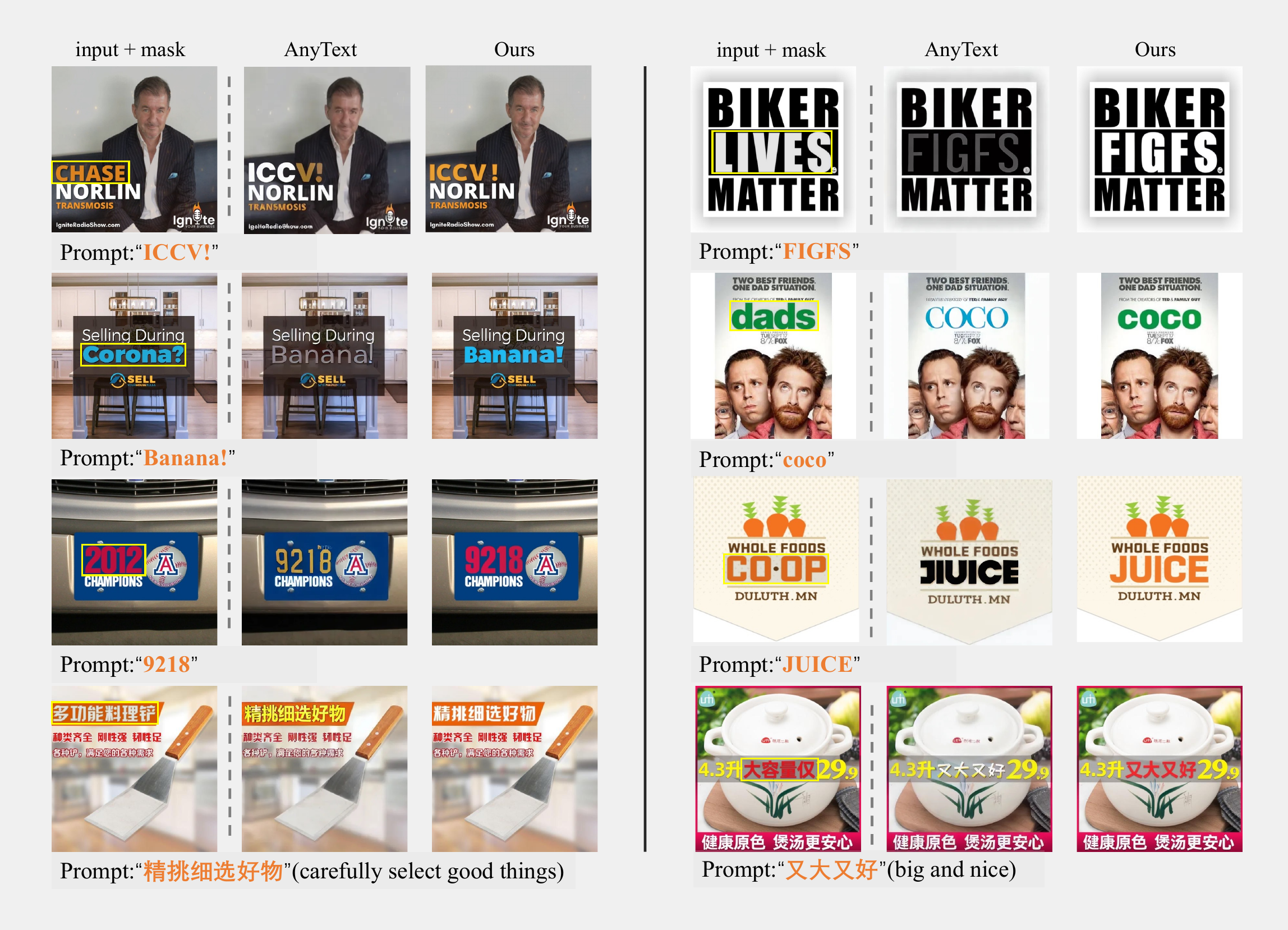}
  \caption{Comparison of results between \textbf{TextMaster} and the state-of-the-art AnyText method. The AnyText inference model utilizes the official open-source version, with test images sourced from the AnyText benchmark dataset.}
  \label{fig:style}  
\end{figure}

\begin{figure*}[ht]
  \centering
  \includegraphics[width=1.0\linewidth]{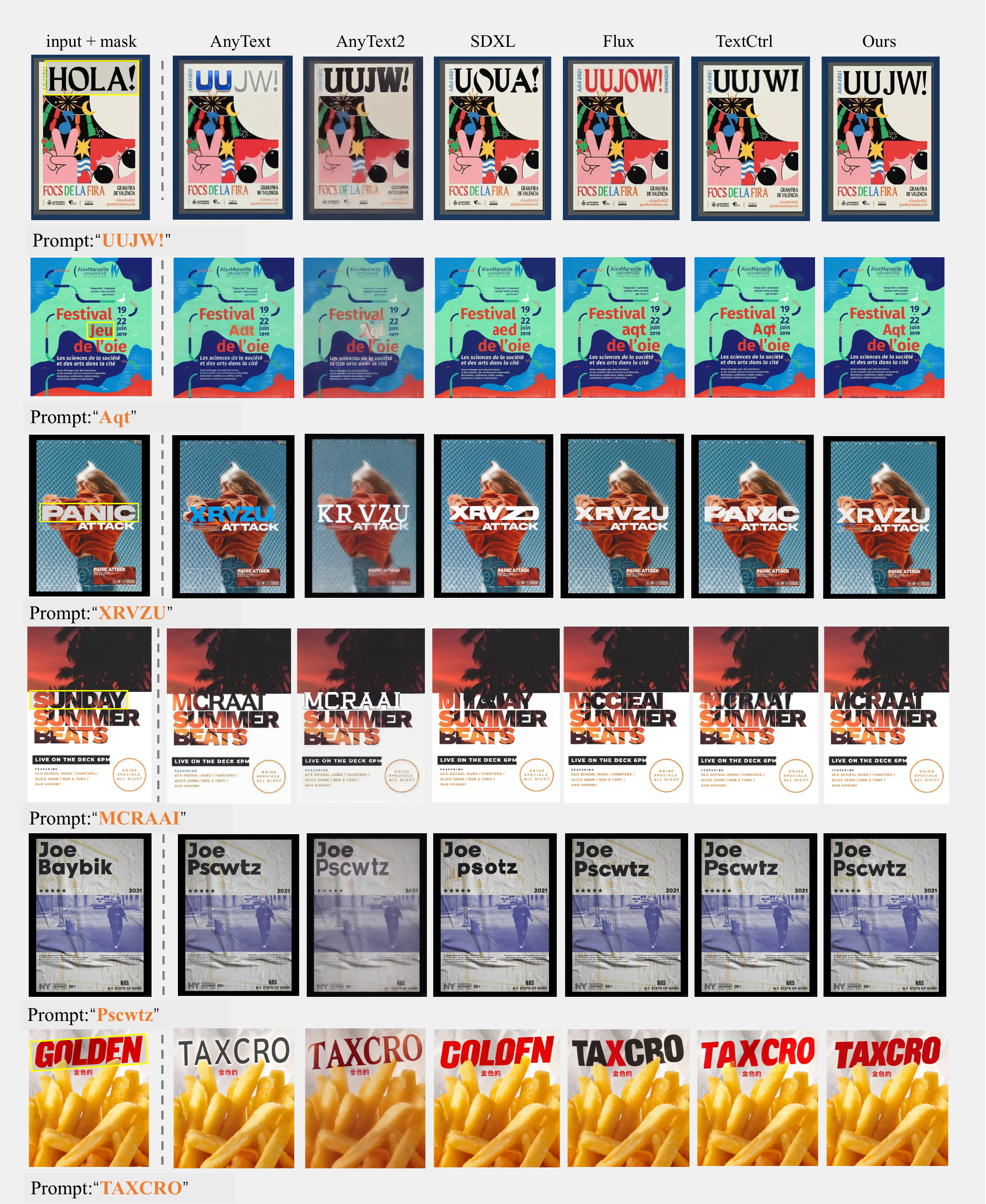}
  \caption{Visual comparison results.}
  \label{fig:compare_all}  
\end{figure*}

\paragraph{Detailed Description of Style Retention Capability.} The re-edited text should maintain the same style as the original, including font style, font color, and font size. However, current text editing methods typically rely on the style of surrounding text, the overall style of the image, or the model's inherent memory to generate style information. As shown in Fig.~\ref{fig:style}, TextMaster seamlessly integrates the original text style into the newly generated text, whereas state-of-the-art methods can only produce styles based on external conditions, often resulting in random style generation. Additionally, these qualitative results demonstrate that TextMaster excels in layout and typesetting capabilities, further highlighting its superiority. More visual comparison results with the full method are presented in Fig.~\ref{fig:compare_all}.

\begin{figure}[ht]
  \centering
  \includegraphics[width=1.0\linewidth]{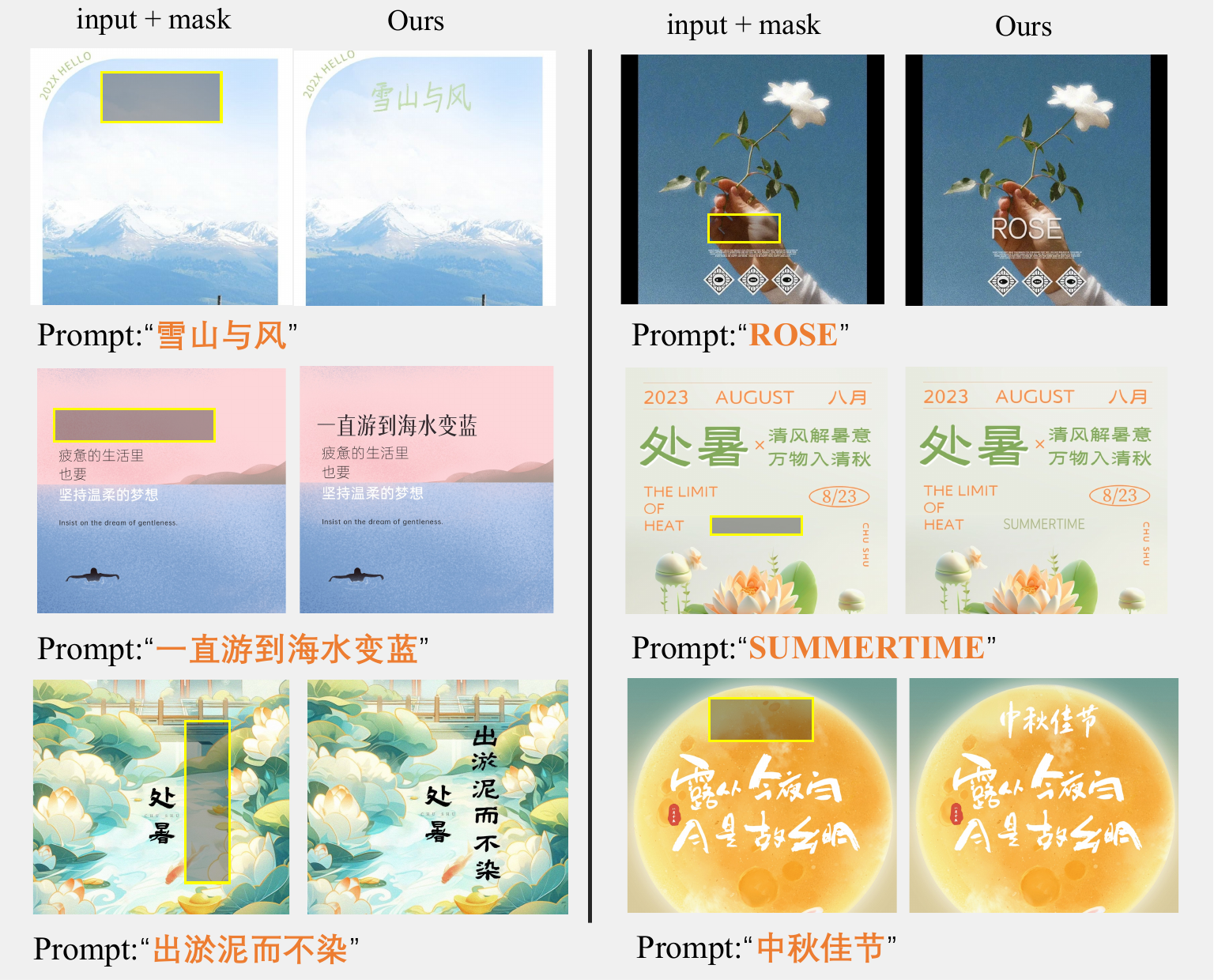}
  \caption{General text rendering visualization.}
  \label{fig:text_render}  
\end{figure}

\paragraph{General Text Rendering Capability.} As shown in the Fig. \ref{fig:text_render}, although our method is not tailored for synthesis scenarios, it can be easily extended to general text rendering. Our method is capable of generating harmonious text rendering effects based on the surrounding background and reference text. Compared to text-to-image-based text rendering approaches, our model offers greater controllability over the placement of the text, preserves the original elements of the image, and maintains high-resolution outputs.

\section*{Limitations and Future Work}
TextMaster currently lacks the capability to simultaneously edit non-continuous multi-line text, which is a challenge we aim to address in future work. Furthermore, beyond simply ensuring that the style of the target text matches that of the original, our future work will focus on enabling the controlled integration of arbitrary styles into the target text.


\end{document}